\documentclass{article}
\usepackage[utf8]{inputenc}
\usepackage{amsmath}
\usepackage{graphicx}
\usepackage{cleveref}
\usepackage{setspace}
\usepackage{natbib}

\title{
Low-memory convolutional neural networks \\ through incremental depth-first processing
}

\author{Jonathan Binas, Yoshua Bengio}
\date{\today}

\begin{document}

\maketitle

\begin{abstract}
We introduce an incremental processing scheme for convolutional neural network (CNN) inference, targeted at embedded applications with limited memory budgets. Instead of processing layers one by one, individual input pixels are propagated through all parts of the network they can influence under the given structural constraints. This depth-first updating scheme comes with hard bounds on the memory footprint: the memory required is constant in the case of 1D input and proportional to the square root of the input dimension in the case of 2D input.
\end{abstract}

\section{Introduction}

Convolutional neural networks (CNNs) deliver state of the art results in most computer vision tasks, such as image segmentation, classification, and object recognition \citep{krizhevsky2012imagenet, lecun2015deep}.
Due to their versatile fields of application, CNN models are being integrated into embedded systems, such as smartphones, internet of things (IoT) endpoints, robotics, or hearing aids.
However, the extensive memory requirements of such models pose serious challenges to the system designer, and have led researchers to explore various approaches for increasing the efficiency of CNN model implementations.
Promising ways of reducing memory requirements and/or lowering the number of operations required to run the network include model compression techniques \citep{ullrich2017soft,louizos2017bayesian,alvarez2017compression,choi2018universal,cho2017mec,han2015deep}, pruning \citep{li2016pruning,lebedev2016fast,yang2017designing,anwar2017structured}, parameter and variable quantization \citep{hubara2016quantized, wu2016quantized,hubara2016binarized,li2016ternary,zhu2016trained,courbariaux2015binaryconnect}, or architectural optimizations \citep{howard2017mobilenets,zhang2017shufflenet}.

In this work, we achieve a substantial reduction in memory occupation while not altering the model itself, but rather the way updates are computed.
Thus, our method is applicable to any CNN model, whereby the computed output is identical to the original output.
In addition to its low memory properties, our technique naturally leads to a sparse update scheme, where no unnecessary computation is carried out if values are zero.
This sparsity, which arises from an event-based processing style employed here, can be exploited, for instance, in conjunction with low-precision parameters and activations, which typically encourage a sparse activation vector.

Unlike other approaches in this domain, we do not provide a method for better parallelization of CNN computation, but rather a streaming core, that can process small segments of the input as they arrive. This is particularly relevant to embedded systems, where memory is scarce and taking into account input sparsity can have a significant impact on the power budget.

\section{Related work}

Depth-first processing schemes for CNNs have been proposed previously. Typically, such methods are based on the identification of computational subgraphs, which can be evaluated independently, leading to a certain level of parallelization.
\citet{alwani2016fused} propose an updating scheme, where a local variable in the network is computed based on all inputs affecting this variable. This leads to multiple evaluations of a given input pattern.
\citet{rouhani2017deep3} map a network to a set of local sub-networks, which can be computed individually.
\citet{weber2018brainslug} focus on accelerating the evaluation of element-wise and pooling layers by identifying independent paths and creating local subgraphs.

Our approach does not rely on advanced methods for finding independent structures, and does not explicitly require independent structures at all.
Instead of asking, which inputs affect a particular output, we ask, which outputs are affected by a particular input.
As a consequence, computations in our model are based on arbitrarily small input segments, allowing for streaming implementations and natural utilization of input sparsity.

\section{Asynchronous incremental updates}

Rather than computing complete feature maps layer by layer, we propose to process inputs in a depth-first fashion.
The advantages of our approach arise from processing small segments, or even single components of the input individually, while keeping in memory only a fraction of the whole network state: only those components of the network state which are non-zero and can be influenced by the current or future input segments (or components) need to be stored.

As an illustration of our method, consider a multilayer neural network (we will add convolutions later), where the activation of a unit $i$ of layer $n$ is given by
\begin{align}
    h_i^{(n)} = f\left(\textstyle\sum_j w_{ij}^{(n)} h_j^{(n-1)} + b_i^{(n)}\right)\,,
    \label{eqn:neuron}
\end{align}
where $f$ is a non-linearity, $w_{ij}$ are the weight parameters, $b_i$ is a bias term, and $h^{(0)}\equiv x$ is the input layer.
Rather than computing whole layers $h^{(1)},\,h^{(2)},\ldots$ one by one for a given input sample, we can process individual components of the input separately.
To do so, we introduce a stateful equivalent of the neuron eq.~\ref{eqn:neuron}, given by
\begin{align}
    \hat h_i^{(n)} = f(c_i^{(n)})\,,
\end{align}
where $c_i^{(n)}$ is a state variable, which is retained throughout the presentation of an input sample.
Furthermore, the neuron emits changes of its activation in terms of ``events'',
\begin{align}
    \Delta \hat h_i^{(n)} = \hat h_i^{(n)}(t) - \hat h_i^{(n)}(t-1)\,,
\end{align}
at times $t$ where such events would be non-zero.
Without loss of generality, we assume the initial state of the network to be zero, $\hat h_i^{(n)}=0\,\forall i,n$, $f(0) = 0$, and $b_i=0\,\forall i$.
An event $\Delta \hat h_j^{(n-1)}$ is transmitted to a connected neuron $i$ via a weight $w_{ij}$ and added to its state variable,
\begin{align}
    c_i^{(n)} \mapsto & c_i^{(n)} + w_{ij}^{(n)}\Delta \hat h_j^{(n-1)},\ \textrm{and thus} \\
    \hat h_i^{(n)} \mapsto & f\left(c_i^{(n)} + w_{ij}^{(n)}\Delta \hat h_j^{(n-1)}\right)
     = \hat h_i^{(n)} + \Delta\hat h_i^{(n)}\,.
\end{align}
Note that a potential change of the target neuron, $\Delta\hat h_i^{(n)}$, is now transmitted in the same way to connected neurons of higher layers.
It is now straight-forward to show that the states $\hat h_i^{(n)}$ and $h_i^{(n)}$ are equal given that both networks have received the same input at their input layers (see \citet{binas2016deep} for details).

\begin{figure}[th]
    \centering
    \includegraphics[width=0.47\textwidth]{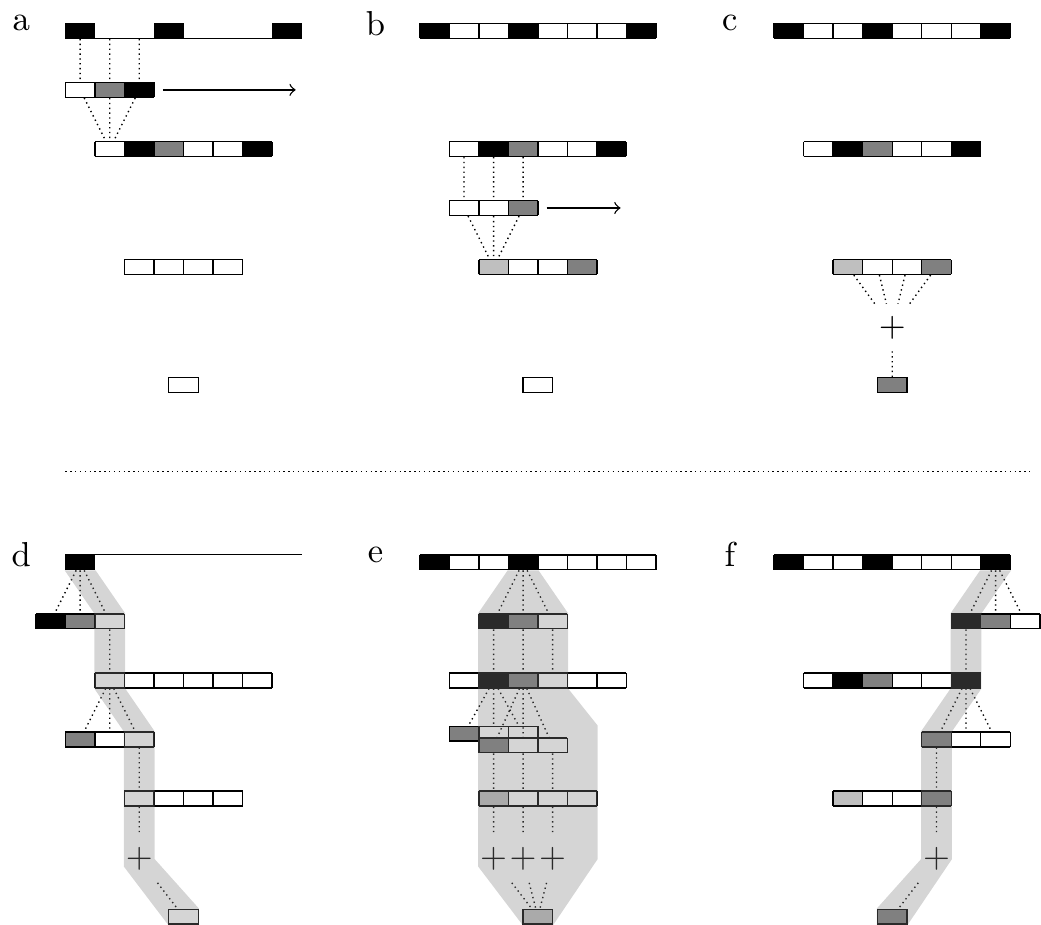}
    \caption{Illustration of the update scheme applied to a one-dimensional model with two hidden layers (each described by a $3 \times 1$ convolutional kernel) and one (averaging) output neuron. The top panels (a-c) show three consecutive update steps of a traditional CNN, where complete layers are computed one by one.
    The bottom panels (d-f) show three steps of the incremental update procedure for the same network, where in each step the effects of one of three non-zero input pixels on the whole network are computed. Only the shaded regions need to be kept in memory.
    Panel e shows the maximum memory that will ever be required by this model (7 activations to be stored,) regardless of the input size.
    %TODO: add more detailed model description.
    }
    \label{fig:illust}
\end{figure}

\begin{figure*}[th]
    \centering
    \includegraphics[width=0.8\textwidth]{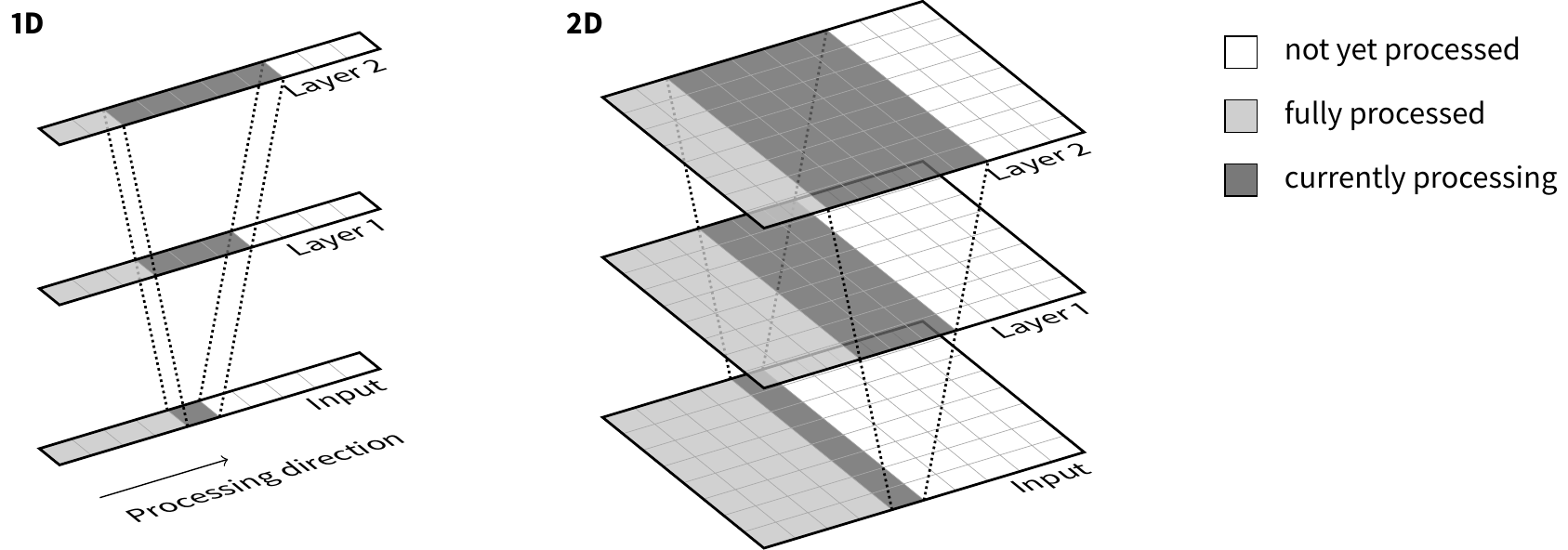}
    \caption{Comparison of the 1D and 2D case. The proposed approach is illustrated in terms of 1D convolutions, but easily scales to higher-dimensional scenarios: in the two-dimensional case, instead of presenting one pixel at a time, a full row/column of an image can be presented and processed at a time. Only the dark shaded regions need to be kept in memory. The white and light shaded regions have not been processed yet, and the light shaded regions have been fully processed and can be forgotten.
    }
    \label{fig:layers}
\end{figure*}

The same event-based update scheme can be applied to CNNs, whereby each neuron only connects to a small number of neurons of the next higher layer, as specified by the convolution kernel.
Thus, each neuron can only influence a few neurons of the next higher layer, typically leading to a cone-shaped dependency structure.
Consequently, a single component of the input only influences a cone-shaped subspace of the whole state space, and thus, it is sufficient to keep only the respective cone in memory when the input component is processed.
In the one-dimensional case, where the input is a vector, consecutive components can be presented one by one, and their contribution to the state cone and the output be computed individually.
Thereby, the memory requirements are constant, regardless of the size of the input vector. This is illustrated in fig.~\ref{fig:illust}.
Conversely, if the input is two-dimensional, a whole row of cones needs to be stored, as their states are still relevant during the presentation of the next row.
Thus, the memory requirements scale with the square-root of the input dimension, in this case (see fig.~\ref{fig:layers} for illustration.)

\section{Experimental verification}

\begin{figure}[th]
    \centering
    \includegraphics[width=0.47\textwidth, trim=0 1cm 0 1cm]{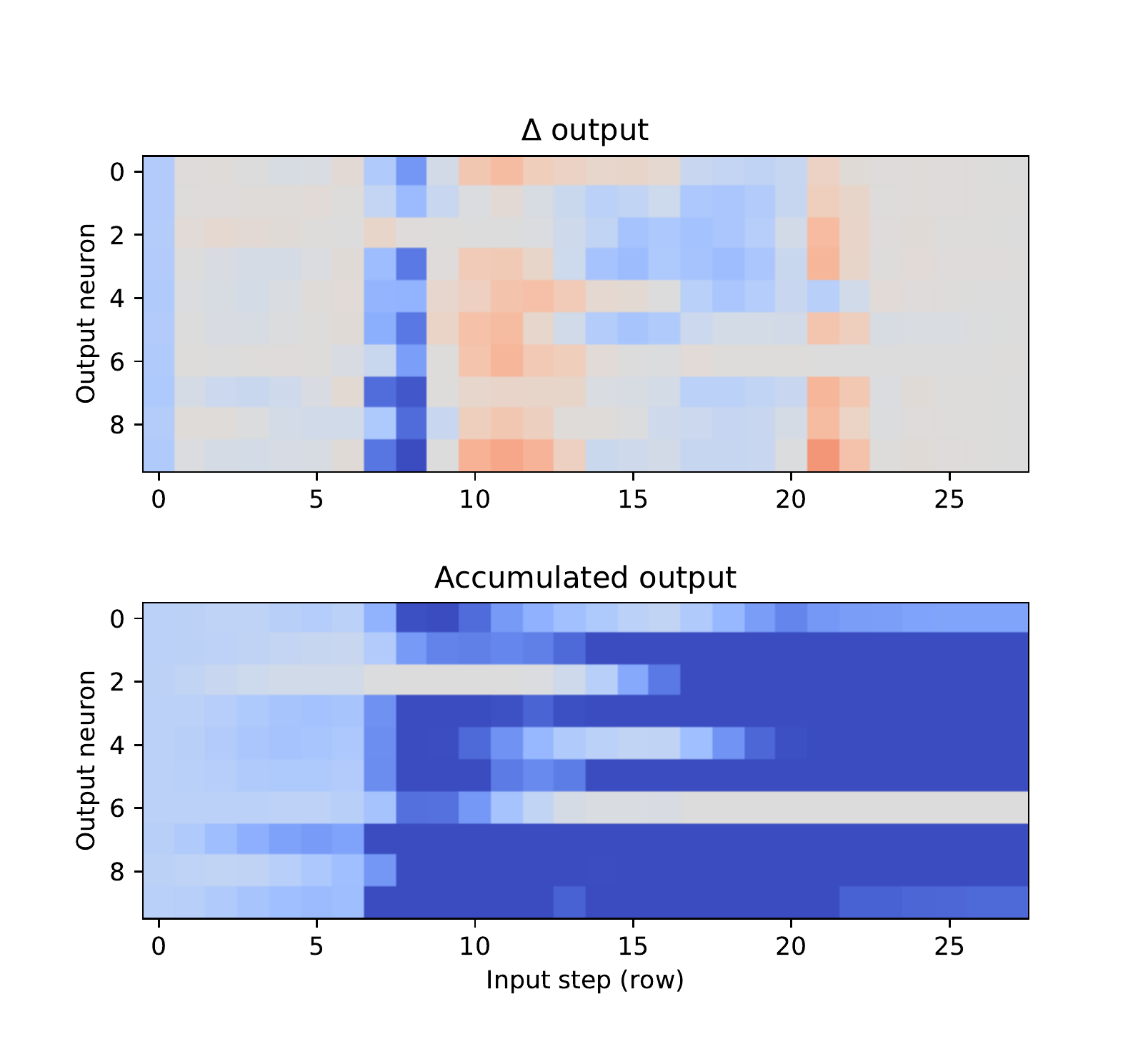}
    \caption{
    Network output over time for an MNIST classifier. The top panel shows the output events over the individual presentation of the 28 rows of an MNIST image. The bottom panel shows the accumulated output (activation of the 10 output units) up to the respective step. The accumulated output converges to the target output.
    Blue means negative, red means positive.
    }
    \label{fig:out}
\end{figure}

\begin{figure}[th]
    \centering
    \includegraphics[width=0.47\textwidth, trim=0 3.5cm 0 0]{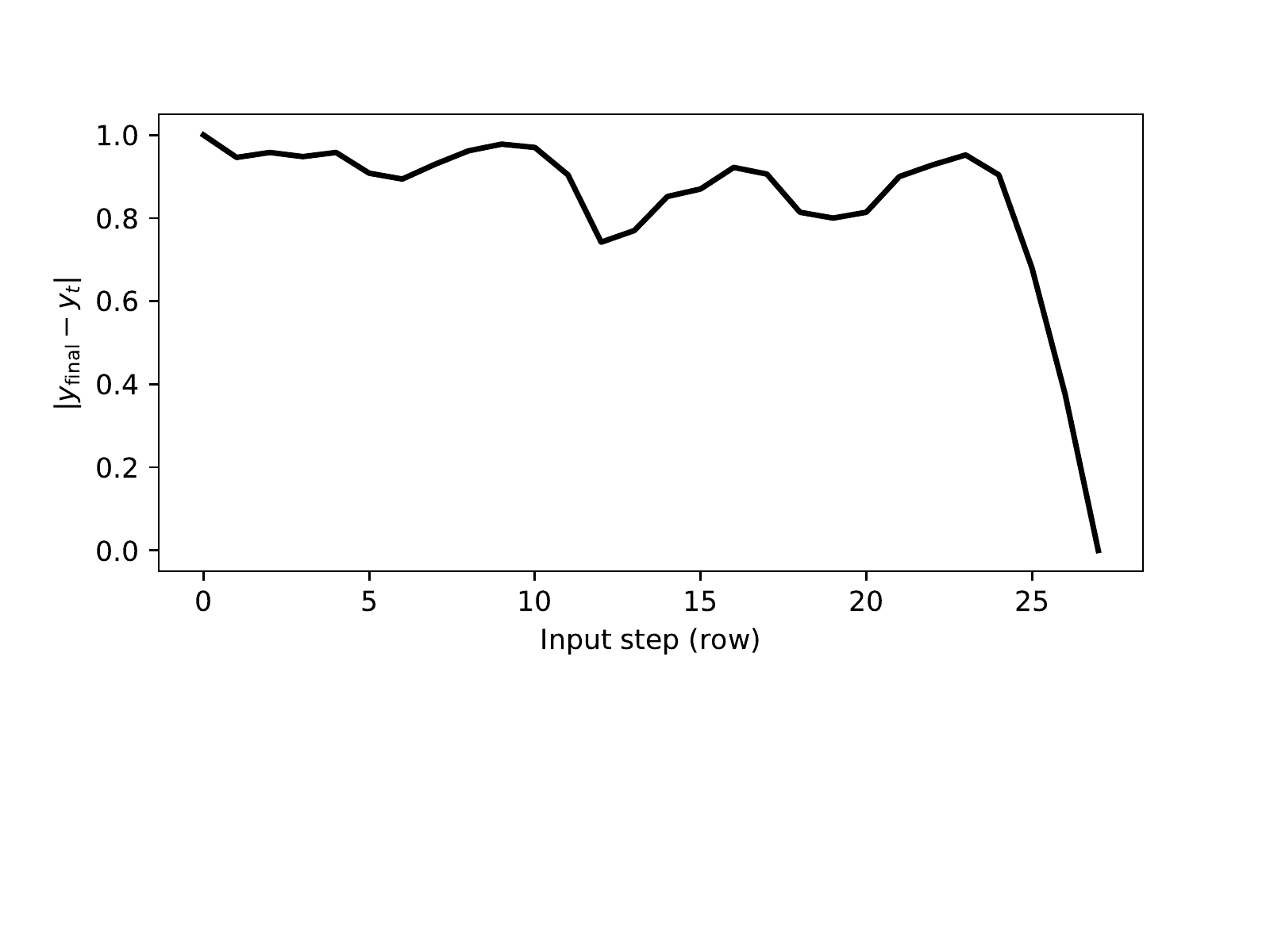}
    \caption{
    Convergence of the output. The figure shows the convergence of the accumulated output to the final output for a classifier trained on MNIST (averaged over 1000 examples.)
    }
    \label{fig:convergence}
\end{figure}

To verify our approach, we implemented the incremental depth-first update mechanism in contemporary deep learning software frameworks\footnote{Code will be made available upon acceptance of the paper.}.
As a proof-of-concept, a multilayer CNN was trained to classify the MNIST dataset. Training was performed using a conventional CNN implementation; the learned parameters were then copied to our incremental model, which was used to classify the dataset.
The two-dimensional input images were fed to the system one row at a time.
Fig.~\ref{fig:out} depicts the output of the network for one example image over the course of the 28 input presentations. The first few inputs (mostly zero, as the digits are roughly centered in the image) have little impact on the output, and the output abruptly changes as more relevant pixels come into view.
The output gradually converges to the exact same prediction value as would be obtained from the underlying conventional CNN.
The convergence speed is shown in fig.~\ref{fig:convergence}. For MNIST, the network seems to require most of the image in order to be able to make an accurate prediction.
It is conceivable that in other scenarios, for instance where the input contains redundant information, better predictions might be obtained earlier during presentation.
The incremental implementation requires only a fraction of the memory of the conventional CNN implementation, as only subsets of the activations need to be retained.
The memory requirements thereby only grow with the square root of the input dimension.

%\section{Hard bounds on memory footprint}
%
%Provide some formal bounds here...
%
%

\section{Discussion}

We propose an incremental updating scheme for convolutional neural networks, where small segments of the input are processed depth-first, and the output layer is gradually updated, as more inputs are presented. With more input segments being presented, the output converges to the same value as a conventional CNN implementation.

Our method puts a hard bound on the memory requirements of the model, and is well suited for embedded applications.
Once implemented, our model operates as a streaming core, taking inputs as they arrive, and incrementally updating its output.

This opens an interesting perspective on CNN processing, which in our framework can be seen as a somewhat recurrent architecture, which has a hidden state (activation subset buffers) and computes an output based on the sequence of inputs (e.g. rows or columns of an image) it receives.
Similar updating schemes as the one proposed here have also been used to implement biologically plausible spike-based learning \citep{o2017temporally}.

Future work will focus on providing a low-level library for embedded systems, as well as the design of network architectures which are particularly suitable for this kind of processing (note that our model does not pose any constraints on the architecture, but it might be more or less efficient depending on the particular network.)

\bibliographystyle{humannat}
\bibliography{bibliography}

\end{document}